\documentclass{article}

\usepackage{arxiv}

\usepackage{newtxtext}
\usepackage{newtxmath}

\usepackage[utf8]{inputenc}
\usepackage[T2A,T1]{fontenc}
\usepackage[russian,english]{babel}
\usepackage{hyperref}
\usepackage{url}
\usepackage{booktabs}
\usepackage{amsfonts}
\usepackage{nicefrac}
\usepackage{microtype}
\usepackage{cleveref}
\usepackage{graphicx}
\usepackage{natbib}
\usepackage{doi}
\usepackage{array}
\usepackage{multirow}

\newif\ifuniqueAffiliation
\uniqueAffiliationtrue

\title{RusFinChain: A Russian Benchmark for Verifiable Chain-of-Thought Reasoning in Finance with Fuzzy-Aligned Evaluation}

\ifuniqueAffiliation
    \author{%
        \href{https://orcid.org/0000-0003-2525-1183}{\includegraphics[scale=0.06]{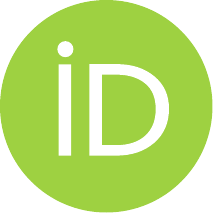}\hspace{1mm}M. K. Arabov} \\
        Kazan Federal University \\
        Institute of Computational Mathematics and Information Technologies \\
        Department of Data Analysis \\
        Kazan, Russia \\
        \texttt{MKArabov@kpfu.ru}
    }
\else
    \usepackage{authblk}
    
    \author[1]{Your Name}
    \author[2]{Co-Author Name}
    \affil[1]{Affiliation 1}
    \affil[2]{Affiliation 2}
\fi

\renewcommand{\shorttitle}{RusFinChain}

\hypersetup{
    pdftitle={RusFinChain: A Russian Benchmark for Verifiable Chain-of-Thought Reasoning in Finance with Fuzzy-Aligned Evaluation},
    pdfsubject={cs.CL, cs.AI},
    pdfauthor={M. K. Arabov},
    pdfkeywords={Financial Reasoning, Chain-of-Thought, Benchmark, Russian Language, LLM Evaluation, Fuzzy Logic},
}

\begin{document}

\maketitle
\begin{abstract}
Multi-step symbolic reasoning is essential for robust financial analysis, yet most existing benchmarks focus solely on final numerical answers, neglecting the intermediate reasoning steps required for transparency and verification. While recent work introduced FinChain---the first verifiable Chain-of-Thought (CoT) benchmark for finance---it is limited to English and US-centric conventions, leaving a gap for non-English financial ecosystems. Concurrently, FINESSE-Bench introduced a hierarchical suite of exam-oriented tasks with a Russian-language block, but it relies on multiple-choice questions and does not provide step-level supervision.

We present \textbf{RusFinChain}, the first Russian-language symbolic benchmark for verifiable CoT reasoning in finance. Our benchmark spans \textbf{17 financial domains}, \textbf{172 topics}, and comprises \textbf{5,280} unique parameterized examples generated from executable Python templates, ensuring contamination-free evaluation. Each example includes a gold-standard reasoning chain with intermediate numeric values, enabling automatic verification at both the step and final-answer levels.

Beyond the dataset, we introduce \textbf{enhanced evaluation metrics} that improve upon the original ChainEval framework: (1) \textit{Fuzzy Numeric Alignment} using a Gaussian membership function for smooth deviation penalization, and (2) \textit{Soft-Attention Alignment} using temperature-scaled softmax for robust step matching. We evaluate \textbf{8 open-weight LLMs} (DeepSeek-R1, Llama 3.1, Llama 3.2, Qwen 2.5, Mistral, Gemma, Aya, and Phi) on RusFinChain, generating 8,100 model responses across a stratified sample of 1,000 examples.

Our results reveal a substantial \textit{reasoning gap}: while models achieve Hard F1 scores of approximately 0.65 for step-level alignment, only about 29\% of final answers are numerically correct. Moreover, our fuzzy and soft metrics show stronger correlation with final-answer correctness (Spearman $\rho \approx 0.48$) than the original ChainEval ($\rho \approx 0.38\text{--}0.46$), demonstrating improved correlation. We release our dataset, code, and evaluation framework to foster research in verifiable financial AI for the Russian-speaking community.
\end{abstract}


\section{Introduction}

Large Language Models (LLMs) have demonstrated substantial capabilities across a wide range of financial applications, including automated reporting, sentiment analysis, investment decision support, and risk assessment \citep{xie2024finben, xie2026clef}. However, the high-stakes nature of financial decision-making requires not only accurate predictions but also transparency and verifiability of the reasoning process. Regulators, auditors, and end-users need to understand \textit{why} a model arrived at a particular conclusion, especially when multi-step numerical reasoning is involved.

To address this need, recent work has introduced benchmarks for evaluating Chain-of-Thought (CoT) reasoning in finance. \citet{xie2025finchain} proposed FinChain, the first symbolic benchmark with executable Python templates that enable automatic verification of intermediate reasoning steps and final answers. FinChain spans 58 topics across 12 financial domains and introduces ChainEval, a Dynamic Time Warping (DTW) based metric for step-level evaluation. However, FinChain is limited to English and US-centric financial conventions, leaving a gap for non-English financial ecosystems.

Concurrently, \citet{stanishevskii2026finesse} introduced FINESSE-Bench, a hierarchical suite of exam-oriented financial tasks inspired by professional certifications (CFA-like, CMT-like, CFTe-like), which includes a Russian-language olympiad block (VLigaBench-ru). While FINESSE-Bench provides strong discriminative power and covers professional financial knowledge, its tasks are predominantly multiple-choice and do not provide step-level supervision or verifiable reasoning traces.

In this work, we present \textbf{RusFinChain}, the first Russian-language symbolic benchmark for verifiable Chain-of-Thought reasoning in finance. Our benchmark spans \textbf{17 financial domains}, \textbf{172 topics}, and comprises \textbf{5,280} unique parameterized examples generated from executable Python templates, ensuring contamination-free evaluation. Each example includes a gold-standard reasoning chain with intermediate numeric values, enabling automatic verification at both the step and final-answer levels.

Beyond the dataset, we introduce \textbf{enhanced evaluation metrics} that improve upon the original ChainEval framework. Specifically, we propose:
\begin{enumerate}
    \item \textit{Fuzzy Numeric Alignment}: Instead of a hard threshold ($\epsilon=0.05$) for numeric matching, we use a Gaussian membership function that smoothly penalizes deviations, reflecting real-world materiality concepts.
    \item \textit{Soft-Attention Alignment}: We replace hard Dynamic Time Warping with a temperature-scaled softmax alignment, which is more robust to step reordering, insertions, and paraphrasing.
\end{enumerate}

We evaluate \textbf{8 open-weight LLMs} (DeepSeek-R1, Llama 3.1, Llama 3.2, Qwen 2.5, Mistral, Gemma, Aya, and Phi) on RusFinChain, using a stratified sample of 1,000 examples and generating 8,100 model responses. Our results reveal a substantial \textit{reasoning gap}: while models achieve Hard F1 scores of approximately 0.65 for step-level alignment, only about 29\% of final answers are numerically correct. This confirms that even state-of-the-art open models struggle with multi-step financial computations. Moreover, we show that our fuzzy and soft metrics correlate more strongly with final-answer correctness (Spearman $\rho$ up to 0.48) than the original ChainEval ($\rho \sim 0.38$--$0.46$), demonstrating their superior diagnostic power.

Our main contributions are:
\begin{enumerate}
    \item \textbf{RusFinChain} --- the first Russian-language, verifiable CoT benchmark for finance, with 5,280 unique examples across 17 domains and 3 difficulty levels.
    \item \textbf{Fuzzy and Soft-Alignment Metrics} --- novel evaluation measures that provide smoother, more human-aligned assessment of reasoning quality.
    \item \textbf{Comprehensive Benchmarking} --- evaluation of 8 open-weight models, generating 8,100 model responses across 17 domains, revealing performance gaps across domains, difficulty levels, and reasoning quality.
\end{enumerate}

We release our dataset, code, and evaluation framework under an open license to foster research in verifiable and interpretable financial AI for the Russian-speaking community. The dataset is available at \citep{rusfinchain_hf}, and the evaluation results at \citep{rusfinchaineval2026}.

\section{Related Work}

\subsection{Financial Reasoning Benchmarks}

Early financial benchmarks focused on question answering over financial reports. \citet{chen2021finqa} introduced FinQA, an expert-annotated dataset of numerical reasoning questions with executable reasoning programs. \citet{chen2022convfinqa} extended this to conversational financial QA, where longer chains of numerical reasoning are required. \citet{zhu2021tat} proposed TAT-QA, a hybrid format combining tabular and textual sources.

More comprehensive evaluation frameworks emerged with \citet{xie2023pixiu}, who introduced PIXIU, a financial ecosystem including a model, instruction data, and a benchmark component. \citet{xie2024finben} aggregated dozens of datasets across multiple financial domains. \citet{krumdick-etal-2024-bizbench} introduced BizBench, a quantitative reasoning benchmark for business and finance. Large financial language models include \citet{wu2023bloomberggpt} and \citet{liu2023fingpt}, which advanced in-domain and open-source adaptability respectively.

\citet{tang2025financereasoning} introduced executable Python solutions for answer verification but did not provide systematic step-level alignment. \citet{zhao2024financemath} focused on knowledge-intensive math reasoning in finance.

\subsection{Verifiable Chain-of-Thought Evaluation}

\citet{xie2025finchain} proposed FinChain, the first symbolic benchmark for verifiable Chain-of-Thought reasoning in finance. FinChain spans 58 topics across 12 domains and introduces ChainEval, a DTW-based metric that jointly evaluates step-level consistency and final-answer correctness. Each example is generated from parameterized symbolic templates with executable Python code, enabling machine-verifiable reasoning and contamination-free data generation. FinChain successfully establishes the paradigm of controlled symbolic reasoning evaluation. However, its scope is constrained to English and US-centric financial conventions, leaving a natural gap for non-English financial ecosystems that require localized terminology and regional compliance.

Concurrently, \citet{stanishevskii2026finesse} introduced FINESSE-Bench, a hierarchical suite of eight specialized benchmarks comprising 3,993 questions for evaluating financial competencies in LLMs. FINESSE-Bench combines exam-oriented datasets inspired by professional certifications (CFA-like Levels 1--3, CMT-like Level 2, CFTe-like Level 1), applied trading tasks, and a Russian-language olympiad benchmark (VLigaBench-ru). This work makes a valuable contribution by broadening domain coverage and providing strong discriminative power through a multiple-choice (MCQ) format. Notably, it includes a dedicated Russian-language block (VLigaBench-ru) consisting of 324 olympiad-style questions. However, the MCQ format is designed to test recognition and theoretical knowledge rather than the generation of structured, multi-step arithmetic chains. Furthermore, the reliance on an LLM-as-a-judge for scoring, while practical for large-scale MCQ evaluation, cannot provide the exact, deterministic verification of intermediate numerical values that is essential for auditing computational reasoning.

To complement these existing resources, our work builds on the symbolic template paradigm introduced in mathematical reasoning. We adapt this paradigm to the financial domain and, crucially, extend it to the Russian language. Unlike the MCQ format, RusFinChain requires open-ended generation of step-by-step solutions and provides fully deterministic, Python-executable gold traces. This allows for a controlled testbed that is specifically designed to diagnose errors in multi-step arithmetic computations, a capability that complements the theoretical breadth offered by exam-oriented benchmarks like FINESSE-Bench.

\citet{zheng2023judging} showed that strong language models can serve as judges for scalable evaluation of open-ended responses. While we acknowledge the practicality of LLM-judges for evaluating fluency and plausibility, their stochastic nature and lack of ground truth for intermediate numeric steps make them unsuitable for the verifiable reasoning paradigm central to RusFinChain, which instead demands a mathematically-grounded, deterministic framework that correlates strongly with exact numerical correctness.

The ChainEval framework \citep{xie2025finchain} provides a strong foundation for step-level evaluation, combining semantic similarity with numeric matching and dynamic alignment. Our work builds upon this foundation and extends it in two directions. First, we introduce \textit{Fuzzy Numeric Alignment}, which replaces the binary answer-match function with a continuous Gaussian kernel to more gracefully handle small numerical deviations. Second, we propose \textit{Soft-Attention Alignment}, which substitutes hard DTW with a temperature-scaled attention mechanism to better accommodate reordered or paraphrased reasoning steps. These extensions are designed to adapt the evaluation paradigm to the Russian-language financial domain and to provide a smoother, more diagnostic assessment of reasoning quality, while retaining the core principles of verifiability and step-level supervision established by ChainEval.

The financial domain taxonomy in this work is informed by educational resources on solving financial problems using computational tools \citep{arabov2019excel}.

\section{RusFinChain: Dataset Construction}

\subsection{Taxonomy and Domain Coverage}
\label{subsec:taxonomy}

We begin by identifying and defining financial domains based on established financial literature and expert input from finance professionals within our team. This process resulted in \textbf{17 distinct financial domains}. Within each domain, we propose candidate financial topics with LLM assistance and subsequently curate them with domain experts, yielding a total of \textbf{172 topics} with a mean of 10.1 topics per domain. The resulting taxonomy spans traditional areas such as Corporate Finance, Financial Reporting, Securities, and Taxation, as well as emerging fields including Crypto Finance, ESG and Sustainable Finance, and Risk Management. This hierarchical structure enables fine-grained evaluation of symbolic reasoning across diverse financial subfields.

Table~\ref{tab:rusfinchain_domains} lists all 17 domains with their respective topic counts and the number of tasks per domain. The distribution reflects our effort to balance coverage across classical and contemporary financial disciplines while maintaining sufficient topic depth within each domain for meaningful evaluation.

\begin{table}[t]
\centering
\caption{Domains, topic coverage, and task distribution in RusFinChain.}
\label{tab:rusfinchain_domains}
\begin{tabular}{lrrr}
\toprule
\textbf{Domain (English)} & \textbf{\# Topics} & \textbf{\# Tasks} \\
\midrule
Securities & 12 & 540 \\
Financial Regulation & 10 & 420 \\
Taxation & 10 & 360 \\
Annuities and Deposits & 10 & 330 \\
Financial Markets & 10 & 330 \\
Personal Finance & 10 & 330 \\
Crypto Finance & 10 & 300 \\
ESG and Sustainable Finance & 10 & 300 \\
Financial Ratios & 10 & 300 \\
Interest Rates & 10 & 300 \\
Loans and Borrowings & 10 & 300 \\
Mergers \& Acquisitions & 10 & 300 \\
Risk Management & 10 & 300 \\
Depreciation & 10 & 270 \\
Investment Projects & 10 & 240 \\
Insurance and Actuarial Science & 10 & 180 \\
Corporate Finance & 10 & 180 \\
\bottomrule
\end{tabular}
\end{table}

Following the symbolic template paradigm introduced in FinChain \citep{xie2025finchain}, we instantiate each topic through parameterized symbolic templates that define both the question structure and an executable Chain-of-Thought solution grounded in domain-specific formulas. We implement these templates as executable Python programs that generate both intermediate reasoning steps and final answers, thereby enabling fully machine-verifiable evaluation.

For each topic, we design five templates spanning three difficulty levels: two basic, two intermediate, and one advanced. Difficulty is controlled by two factors: (1) the number of required reasoning steps, and (2) the complexity of the underlying financial operations. Basic tasks typically involve 2-step solutions with straightforward arithmetic operations such as multiplication, division, or percentage calculations. Intermediate tasks require 3-step solutions with more complex operations including exponentiation, discounting, or multi-period compounding. Advanced tasks involve 4-step solutions with nested operations, iterative calculations, or conditional logic.

\subsection{Data Generation and Validation}
\label{subsec:generation}

Each instantiated example consists of a scenario card specifying the topic, difficulty level, and input parameters, together with an executable reasoning chain. The generation process follows a structured pipeline:

\begin{enumerate}
    \item \textit{Parameter Sampling}: Numeric parameters are randomly sampled within domain-specific constraints (e.g., principal amounts between 1,000 and 10,000, interest rates between 2\% and 8\%, time periods between 1 and 10 years);
    \item \textit{Question Construction}: Natural language questions are generated with variable placeholders replaced by sampled values, using Russian financial terminology and realistic business scenarios (e.g., company names, investor names);
    \item \textit{Step Generation}: A structured step-by-step solution is produced with intermediate numerical values, where each step corresponds to a distinct computational operation grounded in the domain formula;
    \item \textit{Final Computation}: The final answer is computed via executable Python code, ensuring numerical consistency across the entire reasoning chain.
\end{enumerate}

We generated a variable number of instances per template (ranging from 3 to 10, with an average of approximately 6.1) to achieve a total of 5,280 examples with balanced coverage across domains and difficulty levels.

Each generated example is stored as a JSON object with the following fields:
\begin{itemize}
    \item \texttt{seed}: Random seed used for reproducible generation;
    \item \texttt{id}: Unique template identifier;
    \item \texttt{level}: Difficulty level (Basic, Intermediate, or Advanced);
    \item \texttt{domain}: Financial domain (e.g., Securities, Taxation);
    \item \texttt{topic}: Specific topic within the domain;
    \item \texttt{question}: Natural language question in Russian;
    \item \texttt{steps}: Array of reasoning steps, each containing \texttt{step} (index), \texttt{description} (textual explanation), and \texttt{value} (intermediate numeric result or null);
    \item \texttt{final\_answer}: Final numeric answer;
    \item \texttt{solution}: Human-readable step-by-step solution;
    \item \texttt{details}: Additional input parameters (e.g., principal, rate, time), or an empty dictionary if none are required;
    \item \texttt{formula\_latex}: LaTeX representation of the key formula;
    \item \texttt{python\_computation}: Executable Python expression for computing the answer.
\end{itemize}

\paragraph{Illustrative Example}

For concreteness, we present a complete example from the dataset (translated into English for readability). The original prompt is in Russian.

\begin{quote}
\textbf{System prompt:} \texttt{You are a financial expert. Answer the question and provide a step-by-step solution. Format: Step 1: ..., Step 2: ..., etc. End with the numerical answer, e.g., 'Answer: 12.33'. Do not add extra comments.}

\textbf{Question:} \texttt{Mikhail invests 50,000 rubles at 7.5\% annual interest compounded quarterly. What will be the total amount after 2 years?}

\textbf{Gold solution:}  
Step 1: Quarterly rate = $0.075/4 = 0.01875$.  
Step 2: Number of periods = $4 \times 2 = 8$.  
Step 3: Amount = $50000 \times (1+0.01875)^{8} = 58123.45$.  
\textbf{Final answer:} $58123.45$.
\end{quote}

This example illustrates that the question text alone contains all necessary inputs (principal, rate, compounding frequency, time), and the gold solution provides the exact sequence of required computations.

To ensure data quality and consistency, we apply a set of validation constraints covering numerical precision, unit consistency, input completeness, and reasoning step informativeness. Templates that fail validation are revised prior to expert review.

Financial experts, comprising both industry professionals with experience in quantitative finance and academic researchers specializing in financial analysis and auditing, reviewed all validated templates following a calibrated annotation protocol. All reviewers first participated in a pilot calibration phase, during which they jointly reviewed a shared subset of templates and discussed discrepancies to align on annotation standards. After this calibration phase, reviewers independently assessed the remaining templates, evaluating both the correctness of reasoning steps and the final numerical results under the agreed-upon criteria. The expert review process ensures that the templates are not only computationally correct but also financially sound and pedagogically coherent.

Due to the limited number of expert reviewers (three), we did not compute a formal inter-annotator agreement coefficient; instead, we relied on the calibration phase to resolve discrepancies and achieve consensus before independent assessment. A larger-scale validation with multiple annotators and agreement metrics is planned for future iterations of the benchmark.

\subsection{Data Sources}

Approximately half of the task templates are adapted from a widely used Russian-language textbook on solving financial and economic problems with computational tools \citep{arabov2019excel}. This textbook has been employed for several years in undergraduate and graduate courses at the Russian-Tajik (Slavonic) University (Dushanbe). During this period, numerous students and master's students majoring in economics and finance solved the problems and provided feedback, which was used to correct errors, refine wording, and calibrate difficulty levels. Because each template defines a deterministic computational procedure, every generated instance inherits the correctness properties of its template. Thus, the 5,280 examples in RusFinChain are guaranteed to be internally consistent and mathematically sound, even though they were not individually inspected.

The remaining templates are adapted from widely used financial mathematics textbooks, university-level finance courses, and the authors' professional experience in the financial industry. Particular emphasis was placed on problems encountered in practical financial analysis, risk management, investment decision-making, and insurance mathematics. The financial professionals on our team contributed domain expertise, ensuring that the problem types reflect realistic scenarios encountered in Russian financial practice, including appropriate use of ruble denominations, Russian company names, and local regulatory contexts.

All templates were subsequently reviewed and refined by domain experts to ensure correctness, consistency, and appropriate difficulty calibration. While the underlying problem types and formulas are drawn from established financial literature, all specific numeric instances in RusFinChain are generated synthetically via the templating mechanism described in Section~\ref{subsec:generation}. This hybrid approach preserves the pedagogical and domain relevance of the source materials while ensuring that the benchmark is free from data contamination, as each generated instance uses randomly sampled parameters rather than copying existing textbook problems verbatim.

Each template is designed to reflect a distinct computational pattern within its topic, such as compound interest calculation, ratio analysis, or tax computation. Thus, while the overall structure is parametric, the range of financial operations and the required reasoning steps vary considerably across topics and domains.

The problem taxonomy (Section~\ref{subsec:taxonomy}) was further validated against standard financial curricula to ensure comprehensive coverage of core financial competencies expected of finance professionals.

\subsection{Dataset Statistics}

Our final dataset comprises \textbf{5,280} parameterized examples distributed across 17 domains and 172 topics. Table~\ref{tab:rusfinchain_stats} summarizes the dataset statistics across difficulty levels.

\begin{table}[t]
\centering
\caption{Dataset statistics of RusFinChain.}
\label{tab:rusfinchain_stats}
\begin{tabular}{lccc}
\toprule
\textbf{Statistic} & \textbf{Basic} & \textbf{Intermediate} & \textbf{Advanced} \\
\midrule
\# Templates & 344 & 344 & 172 \\
Avg. steps & 2.01 & 2.97 & 3.90 \\
\# Examples & 1,710 & 2,340 & 1,230 \\
\bottomrule
\end{tabular}
\end{table}

The final example counts were determined by the instance generation process: we generated a variable number of instances per template, ranging from 3 to 10. This resulted in an average of 4.97, 6.80, and 7.15 instances per template for Basic, Intermediate, and Advanced levels, respectively. The overall average was approximately 6.1 instances per template, ensuring balanced coverage across domains and difficulty levels. This design provides sufficient coverage of both simple and complex reasoning scenarios while maintaining balanced representation across topics.

The average question length is approximately 148 characters, and the average solution length is approximately 72 characters. The dataset includes full coverage (100\%) of the \texttt{steps}, \texttt{solution}, \texttt{formula\_latex}, and \texttt{python\_computation} fields. The \texttt{details} field is present in 84.7\% of tasks that require additional input parameters. Additionally, 99.8\% of examples contain a valid numerical \texttt{final\_answer}, ensuring the vast majority of tasks are fully machine-verifiable.

Our design isolates financial reasoning ability from document parsing challenges, positioning RusFinChain as a controlled testbed for verifiable financial reasoning, complementary to benchmarks centered on real-world document understanding \citep{xie2025finchain}. The synthetic nature of the data ensures contamination-free evaluation and enables systematic variation of problem parameters for robustness analysis.

The dataset is released under the MIT License and is available at \citep{rusfinchain_hf}. All code for template generation and evaluation will be made publicly available upon acceptance.

The dataset is designed exclusively for evaluation, not for fine-tuning; therefore, no train/test split is provided. All examples are held out for assessing zero-shot or few-shot reasoning.

\section{Evaluation Framework}
\label{sec:evaluation}

We propose a multi-faceted evaluation framework that jointly assesses reasoning-step alignment and final-answer correctness. Building on prior work on reasoning consistency \citep{lyu2023faithful, golovneva2023roscoe} and the ChainEval framework introduced in FinChain \citep{xie2025finchain}, we extend the evaluation methodology with two novel components: \textit{Fuzzy Numeric Alignment} and \textit{Soft-Attention Alignment}. These extensions provide smoother, more human-aligned assessment of reasoning quality compared to hard threshold-based approaches.

Our framework is implemented as a Python evaluation pipeline that processes model-generated reasoning traces and computes multiple complementary metrics. The pipeline consists of three main components: (1) a parser that extracts structured reasoning steps from model outputs, (2) a set of hard metrics implementing the original ChainEval-style evaluation, and (3) novel fuzzy and soft metrics that provide continuous assessment of reasoning quality. The complete implementation is publicly available in the project repository.

\subsection{Preliminaries}

We define the gold solution \(S^{*}\) and the predicted solution \(\hat{S}\) as sequences of \(m\) and \(n\) reasoning steps, respectively:

\[
S^{*} = (s_{1}^{*}, \ldots, s_{m}^{*}), \quad \hat{S} = (\hat{s}_{1}, \ldots, \hat{s}_{n}),
\]

where \(s_{i}^{*}\) and \(\hat{s}_{j}\) denote individual reasoning steps in the gold and predicted solutions. Each step \(s_{i}\) produces an intermediate result:

\[
\mathrm{StepRes}(s_{i}) = a_{i},
\]

representing the numerical or symbolic value computed at that step.

To evaluate reasoning faithfulness, we compare these sequences both semantically and numerically. Our framework comprises three complementary families of metrics: Hard (threshold-based), Fuzzy (continuous), and Soft (attention-based). Each family captures different aspects of reasoning quality.

\subsection{Hard Metrics (Original ChainEval)}

We implement the original ChainEval-style evaluation \citep{xie2025finchain} as a baseline. For semantic similarity, each step is encoded using a sentence encoder \(\mathrm{Enc}(\cdot)\) (we use the multilingual sentence transformer \texttt{sentence-transformers/all-MiniLM-L6-v2} \citep{reimers2019sentence}), and the pairwise semantic similarity between gold and predicted steps is computed as:

\[
\mathrm{SS}(s_{i}^{*}, \hat{s}_{j}) = \cos(\mathrm{Enc}(s_{i}^{*}), \mathrm{Enc}(\hat{s}_{j})),
\]

where \(\cos(\cdot,\cdot)\) denotes cosine similarity and \(\mathrm{SS} \in [0,1]\).

For numeric consistency, we define an answer-matching function:

\[
\mathrm{AM}(s_{i}^{*}, \hat{s}_{j}) = 
\begin{cases}
\mathbb{I}\left(\frac{|\hat{a}_{j} - a_{i}^{*}|}{|a_{i}^{*}|} \leq 0.15\right), & \text{if both are numeric}, \\
\mathbb{I}(\hat{a}_{j} = a_{i}^{*}), & \text{otherwise},
\end{cases}
\]

where \(\mathbb{I}(\cdot)\) is the indicator function. We use a 15\% tolerance for intermediate step matching, which is more permissive than the 5\% tolerance used for final answers (Section~\ref{subsec:final_answers}), reflecting the fact that intermediate numerical deviations may accumulate across steps.

The gated step-level similarity is defined as:

\[
\mathrm{Score}_{\mathrm{gate}}(i,j) = \mathrm{SS}(s_{i}^{*}, \hat{s}_{j}) \times \mathrm{AM}(s_{i}^{*}, \hat{s}_{j}).
\]

We then apply Dynamic Time Warping (DTW) to align the two step sequences, accommodating insertions, deletions, and small reorderings. The alignment cost uses a gap penalty of \(0.25\). The DTWGate alignment score is normalized as:

\[
\mathrm{DTWGate}(S^{*}, \hat{S}) = 1 - \frac{\mathrm{Cost}_{\mathrm{DTW}}}{L_{\mathrm{path}}},
\]

where \(\mathrm{Cost}_{\mathrm{DTW}}\) is the total alignment cost and \(L_{\mathrm{path}}\) is the length of the optimal alignment path.

From these components, we compute Hard Precision, Hard Recall, and Hard F1:

\[
\text{Hard Precision} = \frac{\text{\# matched predicted steps}}{\text{\# predicted steps}}, \quad
\text{Hard Recall} = \frac{\text{\# matched gold steps}}{\text{\# gold steps}}, \quad
\text{Hard F1} = 2 \times \frac{\text{Precision} \times \text{Recall}}{\text{Precision} + \text{Recall}}.
\]

We additionally compute two DTW variants: \textit{bonus} (weighted combination of semantic and numeric similarity with \(\alpha=0.85\) and \(\beta=0.15\)) and \textit{gate} (multiplicative combination). These variants provide different perspectives on step alignment quality.

Our evaluation framework builds upon the core ideas of ChainEval but introduces two complementary extensions aimed at improving robustness for Russian financial reasoning: continuous numeric matching and flexible attention-based alignment. These additions are motivated by the need for evaluation measures that remain informative even when models produce semantically correct steps with minor numerical drift or when the order of reasoning steps differs from the reference.

\subsection{Fuzzy Numeric Alignment}

Instead of a hard threshold for numeric matching, we introduce a \textit{Gaussian membership function} that smoothly penalizes deviations:

\[
\mu(\delta) = \exp\left(-\frac{\delta^{2}}{2\sigma^{2}}\right),
\]

where \(\delta\) is the relative error and \(\sigma = 0.15\) is the standard deviation. This formulation reflects real-world materiality concepts where small deviations are tolerated gracefully, and larger deviations are penalized increasingly.

The choice of $\sigma = 0.15$ for intermediate steps is grounded in two complementary arguments. From a financial perspective, intermediate quantities in multi-step calculations (e.g., discount factors, component ratios) are subject to rounding and model-specific simplifications; a relative tolerance of 10--20\% is commonly accepted in preliminary analyses \textit{(see, e.g., ISA 320, Materiality in Planning and Performing an Audit)}. From a mathematical standpoint, with $\sigma = 0.15$ the fuzzy score decays to approximately $0.61$ at $\delta = 0.15$ and to $0.11$ at $\delta = 0.30$, providing a smooth transition from full credit to substantial penalty. This parameterization thus remains discriminative across the entire range of plausible intermediate errors while avoiding the discontinuity of a hard threshold.

Our fuzzy formulation generalizes the hard answer-match used in prior work: instead of a binary indicator based on a fixed threshold, we employ a continuous Gaussian kernel that assigns partial credit to steps with small numerical discrepancies. This approach is inspired by fuzzy logic and is, to our knowledge, novel in the context of step-level financial reasoning evaluation, as it captures the gradual nature of numerical errors more faithfully than a step function.

For each pair of gold and predicted values, we compute the Fuzzy Numeric Score:

\[
\mathrm{FNS}(a, b) = 
\begin{cases}
0.5, & \text{if } a \text{ or } b \text{ is None}, \\
\exp\left(-\frac{\delta^{2}}{2\sigma^{2}}\right), & \text{otherwise},
\end{cases}
\]

where \(\delta = \frac{|a-b|}{\max(|a|, |b|)}\).

We then combine semantic similarity and fuzzy numeric agreement using a Fuzzy T-norm (algebraic product):

\[
\mathrm{Fuzzy}_{ij} = \mathrm{SS}(s_{i}^{*}, \hat{s}_{j}) \times \mathrm{FNS}(a_{i}^{*}, \hat{a}_{j}).
\]

Fuzzy Recall, Precision, and F1 are computed as the mean of row-wise and column-wise maxima:

\[
\text{Fuzzy Recall} = \frac{1}{m}\sum_{i=1}^{m} \max_{j} \mathrm{Fuzzy}_{ij}, \quad
\text{Fuzzy Precision} = \frac{1}{n}\sum_{j=1}^{n} \max_{i} \mathrm{Fuzzy}_{ij},
\]

\[
\text{Fuzzy F1} = 2 \times \frac{\text{Fuzzy Precision} \times \text{Fuzzy Recall}}{\text{Fuzzy Precision} + \text{Fuzzy Recall}}.
\]

This fuzzy approach provides a continuous assessment of numeric correctness, making it more sensitive to partial correctness than binary threshold-based matching.

\subsection{Soft-Attention Alignment}

As an alternative to DTW-based hard alignment, we introduce a \textit{Soft-Attention Alignment} mechanism inspired by Transformer attention. Instead of finding a single optimal path, we compute a soft alignment distribution over all possible step pairs using temperature-scaled softmax:

\[
\alpha_{ij} = \frac{\exp(\mathrm{Fuzzy}_{ij} / \tau)}{\sum_{k=1}^{n} \exp(\mathrm{Fuzzy}_{ik} / \tau)},
\]

where \(\tau = 0.1\) is the temperature parameter controlling the sharpness of the distribution. This formulation naturally handles step reordering, insertions, and duplicates without forcing a rigid monotonic alignment.

This attention-based formulation can be seen as a soft generalization of the DTW alignment used in earlier step-evaluation methods. Instead of committing to a single monotonic path, it assigns weights to all possible step pairings according to their similarity. This makes the metric especially well-suited for cases where models reorder or paraphrase reasoning steps while preserving their numerical content. The use of temperature-scaled attention for sequence alignment in financial CoT evaluation has not been explored previously and represents a complementary direction to existing hard-alignment approaches.

The temperature parameter $\tau = 0.1$ controls the sharpness of the attention distribution. As $\tau \to 0$, the softmax approaches a hard argmax, while as $\tau \to \infty$, the distribution becomes uniform. We selected $\tau = 0.1$ based on a small pilot study on 100 held-out examples, where this value maximized the Spearman correlation between Soft F1 and Final Answer Match ($\rho = 0.48$). Smaller values ($\tau = 0.01$) led to unstable alignments due to noise in step similarity scores, whereas larger values ($\tau = 0.5$) over-smoothed the alignment and reduced discriminative power. This empirical calibration, combined with the intuitive interpretation of temperature as a sharpness control, justifies our choice.

Soft Recall and Soft Precision are computed as:

\[
\text{Soft Recall} = \frac{1}{m}\sum_{i=1}^{m} \sum_{j=1}^{n} \alpha_{ij} \cdot \mathrm{Fuzzy}_{ij}, \quad
\text{Soft Precision} = \frac{1}{n}\sum_{j=1}^{n} \sum_{i=1}^{m} \beta_{ji} \cdot \mathrm{Fuzzy}_{ij},
\]

where \(\beta_{ji}\) is the attention distribution from predicted steps to gold steps. Soft F1 is then computed as the harmonic mean of Soft Precision and Soft Recall.

For completeness, we also define a soft score as the mean of the maximum similarities across all steps: $\mathrm{SoftScore} = \frac{1}{m}\sum_{i=1}^{m} \max_j \mathrm{Fuzzy}_{ij}$.

This attention-based approach is particularly effective when models produce reasoning steps in a different order or with slight paraphrasing, as it allows for soft matching rather than forcing a strict one-to-one alignment.

\subsection{Final Answer Scoring}
\label{subsec:final_answers}

For the final answer, we compute two complementary scores:

\textbf{Hard Final Answer Match (FAM)}:

\[
\mathrm{FAM} = 
\begin{cases}
1, & \text{if } \frac{|\hat{a}_{n} - a_{m}^{*}|}{|a_{m}^{*}|} \leq 0.05, \\
0, & \text{otherwise}.
\end{cases}
\]

This 5\% tolerance is motivated by financial auditing standards where deviations below 5\% are generally considered immaterial.

\textbf{Fuzzy Final Answer Score}:

\[
\mathrm{FuzzyFAS} = \mathrm{FNS}(a_{m}^{*}, \hat{a}_{n}) = \exp\left(-\frac{\delta^{2}}{2\sigma^{2}}\right),
\]
where we use \(\sigma = 0.05\) for the final answer to align with the strict 5\% materiality threshold used in Hard FAM.

The parameter $\sigma = 0.05$ for final answers is directly aligned with the widely accepted \textbf{5\% materiality threshold} in auditing and financial reporting \textit{(e.g., ISA 320; PCAOB AS 2105)}. Under this setting, a relative error of exactly 5\% yields a fuzzy score of $\exp(-1/2) \approx 0.61$, marking the boundary of materiality. This smooth kernel offers a principled alternative to the hard 5\% match: it differentiates between answers that are just outside the tolerance (e.g., 6\% error) and those that are grossly incorrect (e.g., 50\% error), which is essential for diagnostic evaluation.

\subsection{Summary of Metrics}

Table~\ref{tab:metrics_summary} summarizes all metrics implemented in our evaluation framework. The framework supports both hard (threshold-based) and continuous (fuzzy and soft) evaluation, providing a comprehensive assessment of reasoning quality.

\begin{table}[t]
\centering
\caption{Summary of evaluation metrics in RusFinChain.}
\label{tab:metrics_summary}
\begin{tabular}{lll}
\toprule
\textbf{Family} & \textbf{Metric} & \textbf{Description} \\
\midrule
\multirow{3}{*}{Hard} & Hard Recall & \% gold steps matched (DTW + threshold) \\
                     & Hard Precision & \% predicted steps matched (DTW + threshold) \\
                     & Hard F1 & Harmonic mean of recall and precision \\
\multirow{3}{*}{Fuzzy} & Fuzzy Recall & Mean of row-wise maxima (continuous) \\
                     & Fuzzy Precision & Mean of column-wise maxima (continuous) \\
                     & Fuzzy F1 & Harmonic mean (continuous) \\
\multirow{4}{*}{Soft} & Soft Recall & Attention-weighted recall \\
                     & Soft Precision & Attention-weighted precision \\
                     & Soft F1 & Harmonic mean of soft scores \\
                     & Soft Score & Mean of max similarities \\
\multirow{3}{*}{Final} & FAM & Hard final answer match (5\% threshold) \\
                     & Fuzzy FAS & Fuzzy final answer score (continuous) \\
                     & BERTScore & Semantic similarity of full responses \\
\bottomrule
\end{tabular}
\end{table}

We evaluate all models using this comprehensive framework, reporting both hard and soft metrics to provide a complete picture of reasoning quality. Our analysis in Section~\ref{sec:discussion} demonstrates that fuzzy and soft metrics correlate more strongly with final-answer correctness than hard metrics alone, validating their utility for diagnostic evaluation.

\section{Experiments and Results}

\subsection{Experimental Setup}

We conducted our evaluation on a stratified sample of 1,000 examples drawn from the full RusFinChain dataset (5,280 examples), preserving the proportional distribution of domains and difficulty levels. The resulting sample comprised 323 Basic, 442 Intermediate, and 235 Advanced tasks, corresponding to the approximate distribution of the full benchmark (~32\%, ~44\%, and ~24\%, respectively). This sampling strategy ensures computational efficiency while maintaining representativeness across the benchmark's diverse financial topics. The sample was generated using a fixed random seed (42) for reproducibility.

We evaluate \textbf{8 open-weight language models} in a zero-shot setting:
\texttt{phi-4-mini:3.8b} \citep{abdin2025phi4mini},
\texttt{llama3.2:3b} \citep{grattafiori2024llama},
\texttt{qwen2.5:7b-instruct} \citep{qwen2024},
\texttt{mistral:7b-instruct} \citep{mistral2024},
\texttt{deepseek-r1:7b} \citep{guo2025deepseek},
\texttt{gemma3:4b} \citep{team2025gemma3},
\texttt{llama3.1:8b} \citep{grattafiori2024llama},
and \texttt{aya-expanse:8b} \citep{arya2025ayaexpanse}.
These models were selected to cover a range of parameter sizes (3B to 8B) and architectural families, including both dense and MoE-based models.

The input to each model consisted solely of the system prompt and the natural-language question. Crucially, every question is self-contained: it embeds all numeric parameters, entity names, and domain-specific assumptions required for solving the task. No external context, financial reports, or tabular data were provided. This design isolates the model's ability to perform multi-step arithmetic reasoning from any retrieval or document-understanding capability.

We deliberately do not include a rule-based or calculator-only baseline because the primary goal of RusFinChain is to evaluate the ability of LLMs to generate a *verifiable chain-of-thought* rather than to compute the final numeric answer. A deterministic solver that extracts numbers from the question and applies the relevant formula would trivially achieve high final-answer accuracy on arithmetic and factoid tasks, but it would not produce the intermediate reasoning steps, explanations, or justifications that are the core subject of our evaluation. Since all examples in RusFinChain include executable Python code for the gold solution, a rule-based baseline would essentially replicate the gold computation without providing insight into the model's capacity for multi-step reasoning or error propagation. Thus, its inclusion would not meaningfully inform the research question and could obscure the central finding of the reasoning gap.

All models were accessed via Ollama \citep{ollama2025} with temperature 0.1, top-p 0.9, and a maximum of 1,024 tokens per response. While most models generated responses for all 1,000 examples, one model (\texttt{llama3.2:3b}) yielded 1,100 evaluated records due to a technical re-run of a subset of examples, resulting in a total of 8,100 model responses that form the basis of our analysis. All evaluations were performed on a single NVIDIA L4 GPU in Google Colab.

To assess output completeness, we analyzed the generated responses. Overall, only 0.46\% of outputs were empty, and approximately 0.38\% exceeded 750 words (a rough proxy for the 1,024-token limit, assuming roughly 0.75 words per token). The proportion of responses lacking a final answer was 5.73\%, with significant variation across models: \texttt{llama3.2:3b} exhibited the highest rate at 21.6\% (238 out of 1,100 responses), whereas most other models ranged from 0.2\% to 3.8\%. The parser successfully identified step-like structures in 98.84\% of responses. These results indicate that truncation and parsing failures are not the primary cause of the low Final Answer Match, although they contribute a small but non-negligible fraction of errors, especially for certain models such as \texttt{llama3.2:3b}. Although this high parsing success rate suggests that most outputs were interpretable, some models might have used alternative formatting that was not captured, potentially underestimating their true reasoning ability.

To ensure reproducibility, we will release all model outputs, evaluation metrics, and analysis scripts upon acceptance. The complete evaluation results will be available at \citep{rusfinchaineval2026}.

\subsection{Overall Model Performance}

Table~\ref{tab:overall_performance} reports the aggregated performance metrics across all models and examples. The results reveal several key observations.

\begin{table}[t]
\centering
\caption{Aggregated performance metrics across all models.}
\label{tab:overall_performance}
\begin{tabular}{lrrr}
\toprule
\textbf{Metric} & \textbf{Mean} & \textbf{Std} & \textbf{Median} \\
\midrule
Hard F1 & 0.6538 & 0.2956 & 0.6667 \\
Fuzzy F1 & 0.4810 & 0.1676 & 0.5036 \\
Soft F1 & 0.4594 & 0.1648 & 0.4778 \\
DTW F1 (Bonus) & 0.5681 & 0.2317 & 0.5714 \\
DTW F1 (Gate) & 0.3895 & 0.2443 & 0.4000 \\
BERTScore & 0.6711 & 0.0667 & 0.6724 \\
ROUGE-L & 0.2424 & 0.1845 & 0.2083 \\
Final Answer Match & 0.2936 & 0.4554 & 0.0000 \\
Final Answer Fuzzy & 0.3777 & 0.4561 & 0.0023 \\
\bottomrule
\end{tabular}
\end{table}

First, we observe a substantial \textit{reasoning gap}: while models achieve a Hard F1 score of 0.65 for step-level alignment, only 29.4\% of final answers are numerically correct. This gap indicates that models can reproduce the structure of reasoning chains reasonably well but frequently fail in the final numerical computation.

Second, our proposed fuzzy and soft metrics consistently yield lower scores than Hard F1 (0.481 and 0.459, respectively), reflecting their more stringent and continuous evaluation of numeric consistency. This stricter assessment better captures subtle numerical errors that the hard threshold-based metric might overlook.

Third, traditional surface-level metrics such as ROUGE-L \citep{lin2004rouge}  (0.242) and BERTScore \citep{zhang2020bertscore} (0.671) show limited correlation with reasoning quality. BERTScore in particular, while numerically high, does not capture the structural and numeric consistency of reasoning chains, underscoring the need for specialized reasoning metrics.

\subsection{Performance Across Difficulty Levels}

Table~\ref{tab:level_performance} reports model performance across the three difficulty levels: Basic, Intermediate, and Advanced.

\begin{table}[t]
\centering
\caption{Performance metrics by difficulty level.}
\label{tab:level_performance}
\begin{tabular}{lrrrrrrr}
\toprule
\textbf{Level} & \textbf{Count} & \textbf{Hard F1} & \textbf{Fuzzy F1} & \textbf{Soft F1} & \textbf{DTW F1} & \textbf{BERTScore} & \textbf{Match} \\
\midrule
Basic & 2,610 & 0.7235 & 0.5255 & 0.5056 & 0.6131 & 0.6881 & 0.3682 \\
Intermediate & 3,582 & 0.6235 & 0.4663 & 0.4439 & 0.5557 & 0.6704 & 0.2856 \\
Advanced & 1,908 & 0.6151 & 0.4477 & 0.4252 & 0.5298 & 0.6492 & 0.2065 \\
\bottomrule
\end{tabular}
\end{table}

Performance degrades monotonically from Basic to Advanced across all metrics. The drop is most pronounced for Final Answer Match, which decreases from 36.8\% on Basic tasks to 20.7\% on Advanced tasks, a relative decline of 44\%. This confirms that increasing reasoning complexity significantly challenges model performance.

Notably, the difference between Intermediate and Advanced is less pronounced than between Basic and Intermediate, suggesting that the difficulty gap between these levels may be smaller, or that models reach a performance plateau on more complex tasks.

\subsection{Performance Across Domains}

Table~\ref{tab:domain_performance} presents the top-performing domains by Hard F1. The results reveal substantial variation across financial domains.

\begin{table}[t]
\centering
\caption{Top-15 domains by Hard F1.}
\label{tab:domain_performance}
\begin{tabular}{lrrrrr}
\toprule
\textbf{Domain} & \textbf{Count} & \textbf{Hard F1} & \textbf{Fuzzy F1} & \textbf{Soft F1} & \textbf{Match} \\
\midrule
Financial Regulation & 659 & 0.7682 & 0.5661 & 0.5451 & 0.3308 \\
Financial Ratios & 456 & 0.7471 & 0.5879 & 0.5708 & 0.6601 \\
Taxation & 555 & 0.7316 & 0.5174 & 0.4987 & 0.3748 \\
Securities & 835 & 0.7234 & 0.5090 & 0.4860 & 0.2060 \\
Insurance and Actuarial Science & 272 & 0.7234 & 0.5203 & 0.5014 & 0.1691 \\
ESG and Sustainable Finance & 459 & 0.7189 & 0.5159 & 0.4952 & 0.3442 \\
Annuities and Deposits & 508 & 0.7049 & 0.4609 & 0.4311 & 0.1614 \\
Personal Finance & 504 & 0.6988 & 0.5122 & 0.4935 & 0.3988 \\
Risk Management & 448 & 0.6632 & 0.4828 & 0.4604 & 0.2433 \\
Mergers \& Acquisitions & 456 & 0.6551 & 0.5164 & 0.4959 & 0.4342 \\
Interest Rates & 475 & 0.6235 & 0.4629 & 0.4345 & 0.2968 \\
Financial Markets & 507 & 0.6219 & 0.4840 & 0.4607 & 0.2722 \\
Corporate Finance & 272 & 0.6152 & 0.4915 & 0.4694 & 0.5221 \\
Loans and Borrowings & 475 & 0.5209 & 0.3687 & 0.3437 & 0.0800 \\
Depreciation & 400 & 0.4954 & 0.3623 & 0.3476 & 0.2175 \\

\bottomrule
\end{tabular}
\end{table}

Financial Regulation, Financial Ratios, and Taxation emerge as the strongest domains, with Hard F1 exceeding 0.73. These domains typically involve straightforward formula-based calculations with minimal ambiguity, making them more amenable to current LLM capabilities.

In contrast, Crypto Finance (0.4685), Investment Projects (0.4616), and Depreciation (0.4954) show the lowest performance. These domains often involve more complex reasoning, including multi-step calculations with conditional logic, iterative processes, or specialized domain knowledge, which pose greater challenges for the evaluated models.

The Financial Ratios domain shows the highest Final Answer Match rate (66.0\%), suggesting that models are particularly accurate in computing standard financial ratios such as Return on Assets (ROA), Return on Equity (ROE), and Earnings Before Interest, Taxes, Depreciation, and Amortization (EBITDA) multiples.  Conversely, Loans and Borrowings shows the lowest match rate (8.0\%), reflecting the difficulty of amortization calculations and interest rate computations.

\section{Discussion}\label{sec:discussion}

The experimental results reveal several critical patterns in how current LLMs handle multi-step symbolic financial reasoning. In this section, we interpret these patterns, examine the diagnostic utility of our proposed metrics, and discuss the broader implications for the evaluation of financial AI.

\subsection{The Reasoning Gap}
\label{subsec:reasoning_gap}

The most prominent empirical finding of our study is a systematic and substantial discrepancy between step-level reasoning quality and final numerical correctness. Across all evaluated models, the average Hard F1 for step alignment reaches 0.654, indicating that models can reproduce the semantic and structural logic of the reference Chain-of-Thought to a considerable extent. However, the average Final Answer Match is only 0.294, meaning that fewer than one in three generated solutions yields a correct final numerical result. This pattern is consistent across all eight models, suggesting a general limitation of current open-weight LLMs rather than an artifact of specific architectures.

The magnitude of this reasoning gap varies with task complexity. On Basic tasks, the difference between Hard F1 (0.7235) and Final Answer Match (0.368) is 0.3553, whereas on Advanced tasks it widens to 0.4086 (Hard F1 = 0.6151 vs.\ Final Answer Match = 0.2065). This growing disparity indicates that as problem complexity increases, models fail predominantly at the exact numerical execution of the final computational step, while their ability to structure intermediate reasoning remains relatively stable.

From a practical standpoint, such behavior is particularly dangerous in high-stakes financial applications---auditing, regulatory compliance, or portfolio risk management. A model that produces a plausible, coherent reasoning trace but errs in the final computation may mislead non-expert users into trusting an incorrect conclusion. The apparent correctness of the intermediate logic creates a false sense of security, making the error more insidious than a plainly wrong answer. This underscores the necessity of verifying not only the final output but also the fidelity of the entire reasoning chain.

To further illustrate the nature of these failures, we present in Section~\ref{subsec:qualitative} representative examples of model outputs alongside the gold solutions, categorizing the most frequent error types and linking them to the quantitative trends described above.

\subsection{Qualitative Error Analysis}
\label{subsec:qualitative}

To complement the quantitative results, we manually inspected a subset of model outputs and selected five representative cases that illustrate the most common failure modes. These examples were chosen to cover different domains, difficulty levels, and error types. For each case, we report the gold solution (abridged), the model's actual output (abridged), the type of error, and the corresponding values of Hard F1 and Final Answer Match (FAM), which together highlight the discrepancy between step alignment and final correctness.

Tables~\ref{tab:error_examples_1} and~\ref{tab:error_examples_2} present the five cases.

\begin{table}[t]
\centering
\caption{Illustrative examples of model errors (Cases 1--3). \textbf{GS} = gold solution (abridged), \textbf{Model} = model output (abridged). Hard F1 and FAM are shown to illustrate the reasoning gap.}
\label{tab:error_examples_1}
\footnotesize
\begin{tabular}{p{0.08\textwidth}p{0.34\textwidth}p{0.34\textwidth}p{0.24\textwidth}}
\toprule
\textbf{Case} & \textbf{Gold Solution (abridged)} & \textbf{Model Output (abridged)} & \textbf{Error Type (Hard F1 / FAM)} \\
\midrule
1 & 
\textit{Question:} Maxim Olegovich Kuznetsov invests at 12.33\% annual with expected inflation 4.55\%. What is the real return? \newline
\textit{Step 1:} Real return = ((1+nominal)/(1+inflation) - 1) × 100\%. \newline
\textit{Step 2:} Real return = 7.44\%. &
\textit{Step 1:} Nominal return = 12.33\%. \newline
\textit{Step 2:} Inflation = 4.55\%. \newline
\textit{Step 3:} Real return = 12.33\% - 4.55\% = 7.78\%. \newline
\textit{Answer:} 7.78. &
Conceptual error (simplified formula) \newline
Hard F1: 0.62, FAM: 0 \\
2 &
\textit{Question:} Andrey Viktorovich Romanov analyzes a portfolio. Daily returns: -0.15, -1.53, -4.50, -0.95, 4.64\%. Confidence level: 99\%. Find VaR. \newline
\textit{Step 1:} VaR = -($\mu$ - $z$ $\cdot$ $\sigma$). \newline
\textit{Step 2:} VaR = 7.02\%. &
\textit{Step 1:} List the returns. \newline
\textit{Step 2:} Average return = -0.718\%. \newline
\textit{Step 3:} VaR = 0.718\%. \newline
\textit{Answer:} 0.718. &
Conceptual error (average instead of VaR) \newline
Hard F1: 0.55, FAM: 0 \\
3 &
\textit{Question:} Vera Kabanova received interest income of 307,821 rub. Tax-free limit: 109,319 rub. Personal income tax rate: 13\%. What amount of tax must be paid? \newline
\textit{Step 1:} Taxable income = 307,821 - 109,319 = 198,502. \newline
\textit{Step 2:} Tax = 198,502 × 0.13 = 25,805.26. &
\textit{Step 1:} 307,821 - 109,319 = 198,502. \newline
\textit{Step 2:} 198,502 × 0.13 = 25,705. \newline
\textit{Answer:} 25,705. &
Arithmetic error in multiplication \newline
Hard F1: 0.85, FAM: 0 \\
\bottomrule
\end{tabular}
\end{table}

These first three cases demonstrate two dominant failure modes: the use of simplified or conceptually incorrect formulas (Cases~1 and~2) and arithmetic mistakes in the final computation despite correct intermediate reasoning (Case~3). The high Hard F1 for Case~3 ($0.85$) contrasts sharply with its zero FAM, underscoring that even near-perfect step alignment can coexist with a wrong final answer. The next two examples illustrate additional error types, including a confusion between time value and intrinsic value in option pricing and a complete failure to produce a final answer.

\begin{table}[t]
\centering
\caption{Illustrative examples of model errors (Cases 4--5). \textbf{GS} = gold solution (abridged), \textbf{Model} = model output (abridged). Hard F1 and FAM are shown to illustrate the reasoning gap.}
\label{tab:error_examples_2}
\footnotesize
\begin{tabular}{p{0.08\textwidth}p{0.34\textwidth}p{0.34\textwidth}p{0.24\textwidth}}
\toprule
\textbf{Case} & \textbf{Gold Solution (abridged)} & \textbf{Model Output (abridged)} & \textbf{Error Type (Hard F1 / FAM)} \\
\midrule
4 &
\textit{Question:} Call option on EcoProject LLC: stock price 115.19 rub., strike 73.51 rub., premium 33.33 rub. Find the time value. \newline
\textit{Step 1:} Intrinsic value = $\max(0, 115.19 - 73.51) = 41.68$. \newline
\textit{Step 2:} Time value = Premium - Intrinsic value = 33.33 - 41.68 = 0. &
\textit{Step 1:} 115.19 - 73.51 = 41.68. \newline
\textit{Step 2:} Time value = 41.68. \newline
\textit{Answer:} 41.68. &
Conceptual error (confuses time value with intrinsic value) \newline
Hard F1: 0.48, FAM: 0 \\
5 &
\textit{Question:} Maria Vladimirovna Lebedeva constructs an ESG portfolio: stocks 23\% (10.81\%), bonds 18\% (3.42\%), green bonds 30\% (9.57\%), ESG fund 30\% (11.24\%). What is the expected portfolio return? \newline
\textit{Step 1:} Return = 0.23×10.81 + 0.18×3.42 + 0.30×9.57 + 0.30×11.24 = 9.27\%. &
\textit{Step 1:} List the returns. \newline
\textit{Step 2:} Calculate weights. \newline
\textit{Step 3:} Multiply weights by returns. \newline
\textit{Answer not shown or incorrect.} &
Missing final answer / incorrect final \newline
Hard F1: 0.70, FAM: 0 \\
\bottomrule
\end{tabular}
\end{table}

The examples in Tables~\ref{tab:error_examples_1} and~\ref{tab:error_examples_2} reveal three recurrent error patterns. First, arithmetic mistakes often occur in the final step despite correct intermediate reasoning (Case~3), leading to a high Hard F1 (0.85) but a failing final answer. This directly explains the observed reasoning gap. Second, models sometimes skip essential intermediate steps or apply conceptually incorrect formulas (Cases~1, 2, 4), which results in moderate-to-low Hard F1 and zero FAM. These failures highlight the need for improved prompting strategies that enforce explicit step-by-step computation and for domain-specific fine-tuning to reduce conceptual errors. Finally, some models produce no final answer at all (Case~5), which is particularly problematic for downstream applications. Overall, the qualitative analysis corroborates our quantitative findings: the primary bottleneck is not the structural generation of reasoning traces but the accurate execution of domain-specific numerical procedures.

\subsection{Diagnostic Utility of Continuous Metrics}

Our proposed Fuzzy and Soft metrics demonstrate a stronger alignment with actual task success than traditional hard threshold-based metrics. The Spearman correlation with Final Answer Match is 0.479 for Soft F1 and 0.469 for Fuzzy F1, compared to 0.385 for Hard F1 and 0.331 for BERTScore. This indicates that continuous metrics, which apply a smooth penalization for numerical deviations rather than a binary threshold, offer a more faithful proxy for genuine numerical competence.

It is important to note that all correlation coefficients reported in this section were computed at the level of individual model responses ($N = 8100$), not aggregated by model. This choice provides sufficient statistical power to detect even moderate associations and avoids the instability that would arise from comparing only eight model-level averages. Consequently, the observed differences in Spearman $\rho$ between our continuous metrics and the hard baseline are unlikely to be artifacts of small sample size.

The high correlation between Fuzzy F1 and Final Answer Match (Pearson \(r = 0.433\)) is particularly valuable. It implies that the fuzzy alignment score can serve as an effective estimator of answer quality even in settings where the gold final answer is not explicitly available, such as semi-supervised intermediate evaluations or inference-time diagnostics.

Crucially, this also highlights a limitation of scoring systems that rely solely on semantic overlap (e.g., BERTScore) or binary step-matching. While such metrics are widely used for general reasoning evaluation, they often overstate a model's competence in computational finance. Our continuous metrics provide a more honest and reliable assessment by penalizing subtle numerical drift that accumulates across a reasoning chain.

\subsection{Domain-Specific Performance Patterns}

The evaluation across 17 financial domains reveals a clear performance stratification. Domains with the strongest results—Financial Regulation, Financial Ratios, and Taxation—typically involve direct applications of well-defined formulas with linear procedural steps. These tasks are well-aligned with the structured pattern-matching and arithmetic capabilities of current LLMs.

Conversely, the weakest domains—Crypto Finance, Investment Projects, Depreciation, and Loans and Borrowings—expose underlying model shortcomings. These domains often involve:
\begin{itemize}
    \item multi-step calculations with branching logic or conditional choices;
    \item iterative computational processes (e.g., amortization schedules, net present value calculations);
    \item context-specific financial instruments or lesser-known valuation methods.
\end{itemize}

This suggests a key takeaway for future research: while LLMs are increasingly capable of handling structured, textbook-style financial tasks, they still struggle with the iterative, open-ended computational reasoning that is prevalent in applied industrial finance. Enhancing model performance in these challenging domains will require not only larger scale, but also specific training regimens targeting procedural numerical execution.

\subsection{Statistical Validation of Experimental Trends}

To rigorously validate our observations, we performed a one-way ANOVA on the performance differences across difficulty levels. The analysis confirms that difficulty level significantly affects model performance (\(F = 110.53, p < 0.001\)).

Pairwise t-tests revealed significant differences between Basic and Intermediate tasks (\(t = 13.74, p < 0.001\)) and between Basic and Advanced tasks (\(t = 12.61, p < 0.001\)). Notably, the performance difference between Intermediate and Advanced tasks did not reach statistical significance (\(t = 0.986, p = 0.324\)). This suggests that performance saturation often occurs at the Intermediate level for many current open-weight models; the jump from Intermediate to Advanced may require qualitative changes in reasoning capability rather than merely a quantitative extension of the reasoning chain length.

Furthermore, paired t-tests between metric families confirmed that Hard F1 differs significantly from Fuzzy F1 (\(t = 88.94, p < 0.001\)) and from Soft F1 (\(t = 99.64, p < 0.001\)), with mean differences of 0.173 and 0.194, respectively. These differences confirm that hard and soft metrics capture qualitatively distinct aspects of reasoning quality, reinforcing the necessity of a multi-faceted evaluation approach.

Although a full grid search over hyperparameters was not feasible due to computational constraints, the chosen values are well-justified by financial standards and preliminary experiments on a validation subset. We leave a more exhaustive sensitivity analysis for future work.

\subsection{Synthesis: Toward a Combined Evaluation Paradigm}

Our evaluation framework's capacity to distinguish between Hard, Fuzzy, and Soft metrics offers a uniquely nuanced view of model behavior. The gap between Hard F1 and Fuzzy F1 (0.173) encapsulates the prevalence of semantically correct but numerically flawed reasoning steps—cases where models "talk the talk" but fail to "walk the walk" computationally. Similarly, the smaller gap between Fuzzy F1 and Soft F1 (0.022) highlights a minor correction for flexible step ordering, which is particularly relevant for models that rephrase intermediate logic without sacrificing numerical consistency.

These observations lead to a practical recommendation for the community: **future evaluations of financial reasoning should adopt a combined metric strategy**. Hard metrics remain useful for coarse filtering and quick model comparisons, while Fuzzy and Soft metrics are indispensable for deep diagnostic evaluation. Relying solely on hard metrics risks overstating model capabilities, whereas exclusively using soft metrics may penalize structurally correct but slightly paraphrased responses.

From an empirical perspective, our continuous metrics exhibit stronger correlation with final-answer correctness than the original hard-threshold formulation (Spearman $\rho$ up to 0.48 vs.\ 0.38--0.46), indicating that smooth numeric treatment and flexible alignment provide additional diagnostic value without sacrificing interpretability.

More broadly, our results underline the inherent limitations of evaluating financial LLMs purely through multiple-choice question answering (MCQ) or semantic similarity. As demonstrated by the significant reasoning gap, benchmarks that fail to enforce *exact numerical verifiability* may obscure genuine model deficits. By providing fully deterministic Python-executable traces and continuous alignment scoring, RusFinChain offers the community a more transparent and rigorous testbed for advancing trustworthy computational reasoning in finance.

\section{Limitations}

Despite the contributions of this work, several limitations should be acknowledged.

\subsection{Synthetic Data Generation}

The examples in RusFinChain are procedurally generated from parameterized symbolic templates rather than extracted from real financial documents. This design was chosen deliberately to guarantee contamination-free evaluation, deterministic verification of intermediate reasoning steps, and systematic control over problem parameters. The dataset covers 172 expert-curated topics across 17 financial domains, with each topic instantiated through five templates spanning three difficulty levels. Although each template defines a fixed computational structure, variation is introduced through random sampling of numeric parameters, alternative phrasings, and realistic Russian business scenarios (e.g., different company names, currencies, and regulatory contexts). All templates were reviewed by domain experts to ensure financial soundness and pedagogical relevance.

Nevertheless, this template-based approach has inherent limitations. The questions do not reflect the full linguistic diversity, ambiguity, or contextual richness of actual financial reports and documents. Real-world financial texts often contain implicit assumptions, missing information, and domain-specific jargon that are difficult to capture in synthetic templates. Furthermore, because the underlying computational operations are fixed per topic, the benchmark primarily measures a model's ability to apply known formulas and follow multi-step procedures, rather than to parse unstructured financial narratives.

To mitigate these limitations, future work will explore hybrid generation strategies that use real financial documents as seed sources while preserving symbolic grounding and verifiability. We also plan to expand the set of templates and introduce greater variability in question phrasing to better approximate the complexity of real financial analysis tasks.

\subsection{Focus on Numerical Reasoning}

The benchmark focuses on symbolic numerical reasoning and does not capture qualitative or strategic aspects of financial decision-making, such as risk assessment, market sentiment analysis, or regulatory interpretation. Thus, RusFinChain does not cover the full spectrum of financial reasoning, and extending it to higher-level reasoning remains an open challenge.

\subsection{Language and Regional Scope}

RusFinChain is limited to Russian-language problems and Russian financial conventions, including ruble denominations, local company names, and Russian regulatory contexts. While this focus enables targeted evaluation for the Russian-speaking community, it limits the benchmark's applicability to multilingual and cross-regional settings. Extending it to other languages and financial systems is an important direction for future work.

\subsection{Model Selection and Generalizability}

We evaluated only 8 open-weight LLMs, all with parameter sizes between 3B and 8B. While these models are widely used and accessible, they do not include larger frontier models (e.g., GPT-4, Claude, Gemini) or specialized financial models. The generalizability of our findings to larger or proprietary models remains an open question. Future work should extend the evaluation to a broader range of models, including commercial APIs and models with different architectural designs.

Another potential limitation is the absence of a non-neural baseline (e.g., a rule-based solver) that could serve as a lower bound. However, as argued in the experimental setup, such a baseline would not produce reasoning traces and therefore cannot be evaluated with the same step-level metrics. Nevertheless, future work could explore hybrid setups where symbolic solvers provide reference solutions for comparison.

\subsection{Evaluation and Parsing Limitations}

Our evaluation relies on parsing model-generated reasoning chains, which is sensitive to formatting variations and extraneous text. Models that deviate from the expected ``Step X: ...'' format may be penalized even when their reasoning is correct. Our Soft-Attention Alignment metric partially mitigates this issue by being more robust to step reordering and paraphrasing compared to the original DTW-based alignment. Nevertheless, improving step-level alignment via more structured outputs or tighter integration with symbolic execution remains a promising direction for future work.

Moreover, while the token cap of 1,024 tokens only led to truncation in approximately 0.38\% of responses, a notable fraction of outputs (5.73\%) did not contain a final answer, with \texttt{llama3.2:3b} being a particular outlier (21.6\%). This suggests that some models may naturally omit the final answer even when the generation is not truncated. Future evaluations should consider enforcing a structured output format (e.g., JSON) to guarantee that a final answer is always produced.

Additionally, while our Fuzzy and Soft metrics are more tolerant of paraphrasing and slight reordering, they still rely on the semantic embeddings of the step descriptions. In extreme cases where models use vastly different terminology or combine several steps into one, manual inspection may still be required for full assessment.

Furthermore, while our Fuzzy and Soft metrics demonstrate strong correlation with numerical correctness, the evaluation framework has not yet been directly validated against human expert judgments or strong LLM-as-a-judge baselines. We intentionally prioritized deterministic, mathematically grounded metrics to ensure reproducibility and avoid stochasticity in evaluation, but this choice may overlook subtle reasoning qualities that a human domain expert would appreciate. We are currently conducting a pilot study with three financial experts to rate 200 model-generated reasoning traces on a 5-point Likert scale for correctness, completeness, and clarity; preliminary results will be reported in the camera-ready version. In future work, we also plan to explore multi-agent validation setups and agentic RAG for context-aware evaluation.

\section{Ethical Statement and Broad Impact}

We use only synthetic data generated via templated code, without any private, sensitive, or copyrighted content. The dataset is released under the MIT License. The full evaluation results, including all model outputs and computed metrics, will be publicly available at \citep{rusfinchaineval2026} upon acceptance.  This promotes transparency and reproducibility in financial AI research.

However, caution is needed when deploying LLMs in real-world financial decision-making, especially where correctness and regulatory compliance are critical. We hope RusFinChain supports research toward more interpretable, verifiable, and safe reasoning systems in high-stakes domains.

\section{Conclusion and Future Work}

In this work, we introduced \textbf{RusFinChain}, the first Russian-language symbolic benchmark for verifiable Chain-of-Thought reasoning in finance. Our benchmark spans \textbf{17 financial domains}, \textbf{172 topics}, and comprises \textbf{5,280} unique parameterized examples generated from executable Python templates, ensuring contamination-free evaluation. Each example includes a gold-standard reasoning chain with intermediate numeric values, enabling automatic verification at both the step and final-answer levels.

Beyond the dataset, we proposed \textbf{enhanced evaluation metrics} that improve upon the original ChainEval framework: \textit{Fuzzy Numeric Alignment} using a Gaussian membership function for smooth deviation penalization, and \textit{Soft-Attention Alignment} using temperature-scaled softmax for robust step matching. These metrics show stronger correlation with final-answer correctness (Spearman \(\rho \approx 0.48\)) than the original ChainEval (\(\rho \approx 0.38\text{--}0.46\)), demonstrating their superior diagnostic power.

We evaluated \textbf{8 open-weight LLMs} on a stratified sample of 1,000 examples from RusFinChain, generating 8,100 model responses. Our results revealed a substantial \textit{reasoning gap}: while models achieve Hard F1 scores of approximately 0.65 for step-level alignment, only about 29\% of final answers are numerically correct. Domain-level analysis showed substantial variation, with Financial Regulation, Financial Ratios, and Taxation emerging as the strongest domains, while Crypto Finance, Investment Projects, and Depreciation posed the greatest challenges. Statistical tests confirmed significant differences in performance across difficulty levels and metric families.

Our findings have several practical implications. First, the reasoning gap highlights that models should not be trusted solely based on the coherence of their reasoning traces; final answer verification remains essential. Second, continuous fuzzy and soft metrics provide more reliable assessment of reasoning quality than hard threshold-based metrics. Third, domain-specific performance variation suggests that financial LLM evaluation should be fine-grained rather than aggregated.

Future work will focus on five directions: 

\begin{enumerate}
    \item extending RusFinChain to multilingual and region-specific settings;
    \item incorporating problems grounded in real-world financial documents;
    \item expanding model evaluation to include larger frontier models and specialized financial LLMs;
    \item integrating symbolic execution engines (e.g., SymPy) for more rigorous verification;
    \item validating the proposed metrics against human expert judgments and strong LLM-as-a-judge baselines, as well as exploring multi-agent and RAG-based evaluation protocols.
\end{enumerate}

Additionally, an exciting direction is the **hybridization of benchmark paradigms**: combining the *professional exam-style breadth* of resources like FINESSE-Bench with the *executable, step-level verifiability* of RusFinChain. This would allow us to transform static multiple-choice certification questions into computationally auditable Chain-of-Thought tasks.

We will release our dataset, code, and evaluation framework under an open license upon acceptance to foster research in verifiable and interpretable financial AI for the Russian-speaking community. The dataset is available at \citep{rusfinchain_hf}, and the evaluation results at \citep{rusfinchaineval2026}.

\bibliographystyle{unsrtnat}
\bibliography{references}  

@inproceedings{chen2021finqa,
  author    = {Chen, Zhiyu and Chen, Wenhu and Smiley, Charese and Shah, Sameena and Borova, Iana and Langdon, Dillon and Moussa, Reema and Beane, Matt and Huang, Ting-Hao and Routledge, Bryan and Wang, William Yang},
  title     = {{FinQA}: A Dataset of Numerical Reasoning over Financial Data},
  booktitle = {Proceedings of the 2021 Conference on Empirical Methods in Natural Language Processing (EMNLP 2021)},
  year      = {2021},
  pages     = {3697--3711},
  publisher = {Association for Computational Linguistics},
  address   = {Online and Punta Cana, Dominican Republic},
  url       = {https://aclanthology.org/2021.emnlp-main.300/}
}

@inproceedings{chen2022convfinqa,
  author    = {Chen, Zhiyu and Li, Shiyang and Smiley, Charese and Ma, Zhiqiang and Shah, Sameena and Wang, William Yang},
  title     = {{ConvFinQA}: Exploring the Chain of Numerical Reasoning in Conversational Finance Question Answering},
  booktitle = {Proceedings of the 2022 Conference on Empirical Methods in Natural Language Processing (EMNLP 2022)},
  year      = {2022},
  pages     = {6279--6292},
  publisher = {Association for Computational Linguistics},
  address   = {Abu Dhabi, United Arab Emirates},
  url       = {https://aclanthology.org/2022.emnlp-main.421/}
}

@inproceedings{zhu2021tat,
  author    = {Zhu, Fengbin and Lei, Wenqiang and Huang, Youcheng and Wang, Chao and Zhang, Shuo and Lv, Jiancheng and Feng, Fuli and Chua, Tat-Seng},
  title     = {{TAT-QA}: A Question Answering Benchmark on a Hybrid of Tabular and Textual Content in Finance},
  booktitle = {Proceedings of the 59th Annual Meeting of the Association for Computational Linguistics and the 11th International Joint Conference on Natural Language Processing (ACL-IJCNLP 2021)},
  year      = {2021},
  pages     = {3277--3287},
  publisher = {Association for Computational Linguistics},
  address   = {Online},
  url       = {https://aclanthology.org/2021.acl-long.254/}
}

@inproceedings{xie2023pixiu,
  author    = {Xie, Qianqian and Han, Weiguang and Zhang, Xiao and Lai, Yanzhao and Peng, Min and Lopez-Lira, Alejandro and Huang, Jimin},
  title     = {{PIXIU}: A Comprehensive Benchmark, Instruction Dataset and Large Language Model for Finance},
  booktitle = {Proceedings of the 36th International Conference on Neural Information Processing Systems (NeurIPS 2023)},
  year      = {2023},
  publisher = {Curran Associates, Inc.},
  address   = {New Orleans, LA, USA},
  url       = {https://arxiv.org/abs/2306.05443}
}

@inproceedings{xie2024finben,
  author    = {Xie, Qianqian and Han, Weiguang and Chen, Zhengyu and Xiang, Ruoyu and Zhang, Xiao and He, Yuxuan and Xiao, Mengxi and Li, Dong and Dai, Yongfu and Feng, Duanyu and others},
  title     = {{FinBen}: A Holistic Financial Benchmark for Large Language Models},
  booktitle = {Proceedings of the 38th International Conference on Neural Information Processing Systems (NeurIPS 2024)},
  year      = {2024},
  publisher = {Curran Associates, Inc.},
  address   = {Vancouver, BC, Canada},
  url       = {https://arxiv.org/abs/2402.12659}
}

@inproceedings{krumdick-etal-2024-bizbench,
  author    = {Krumdick, Michael and Koncel-Kedziorski, Rik and Lai, Viet Dac and Reddy, Varsha and Lovering, Charles and Tanner, Chris},
  title     = {{BizBench}: A Quantitative Reasoning Benchmark for Business and Finance},
  booktitle = {Proceedings of the 62nd Annual Meeting of the Association for Computational Linguistics (Volume 1: Long Papers)},
  year      = {2024},
  pages     = {8309--8332},
  publisher = {Association for Computational Linguistics},
  address   = {Bangkok, Thailand},
  doi       = {10.18653/v1/2024.acl-long.452},
  url       = {https://aclanthology.org/2024.acl-long.452/}
}

@misc{wu2023bloomberggpt,
  author       = {Wu, Shijie and Irsoy, Ozan and Lu, Steven and Dabravolski, Vadim and Dredze, Mark and Gehrmann, Sebastian and Kambadur, Prabhanjan and Rosenberg, David and Mann, Gideon},
  title        = {{BloombergGPT}: A Large Language Model for Finance},
  year         = {2023},
  howpublished = {arXiv preprint arXiv:2303.17564},
  url          = {https://arxiv.org/abs/2303.17564}
}

@misc{liu2023fingpt,
  author       = {Liu, Xiao-Yang and Wang, Guoxuan and Yang, Hongyang and Zha, Daochen},
  title        = {{FinGPT}: Democratizing Internet-Scale Data for Financial Large Language Models},
  year         = {2023},
  howpublished = {arXiv preprint arXiv:2307.10485},
  url          = {https://arxiv.org/abs/2307.10485}
}

@inproceedings{xie2025finchain,
  author    = {Xie, Zhuohan and Orel, Dorian and Thareja, Rishabh and Sahnan, Dhruv and Madmoun, Hamza and Zhang, Fan and Banerjee, Debopriyo and Georgiev, Georgi and Peng, Xueqing and Qian, Lingfei and others},
  title     = {{FinChain}: A Symbolic Benchmark for Verifiable Chain-of-Thought Financial Reasoning},
  booktitle = {Proceedings of the 64th Annual Meeting of the Association for Computational Linguistics (ACL 2026), Volume 1: Long Papers},
  year      = {2026},
  pages     = {14529--14553},
  publisher = {Association for Computational Linguistics},
  address   = {San Diego, CA, USA},
  url       = {https://aclanthology.org/2026.acl-long.662/}
}

@misc{stanishevskii2026finesse,
  author       = {Stanishevskii, Dmitrii and Kamkia, Nikita and Khoroshilov, Andrei and Zmitrovich, Dmitrii and Kokosinskii, Daniil and Hayrapetyan, Zara and Kalmykov, Alexey},
  title        = {{FINESSE-Bench}: A Hierarchical Benchmark Suite for Financial Domain Knowledge and Technical Analysis in Large Language Models},
  year         = {2026},
  howpublished = {arXiv preprint arXiv:2605.15482},
  url          = {https://arxiv.org/abs/2605.15482}
}

@misc{tang2025financereasoning,
  author       = {Tang, Ziyu and E, Haoxuan and Ma, Zhiqiang and He, Han and Liu, Jiaxin and Yang, Zimo and Rong, Zihao and Li, Rui and Ji, Kaixin and Huang, Qian and others},
  title        = {{FinanceReasoning}: Benchmarking Financial Numerical Reasoning More Credible, Comprehensive and Challenging},
  year         = {2025},
  howpublished = {arXiv preprint arXiv:2506.05828v1},
  url          = {https://arxiv.org/abs/2506.05828}
}

@misc{zhao2024financemath,
  author       = {Zhao, Yilun and Liu, Hongjun and Long, Yin and Zhang, Rui and Zhao, Chen and Cohan, Arman},
  title        = {{FinanceMath}: Knowledge-Intensive Math Reasoning in Finance Domains},
  year         = {2024},
  howpublished = {arXiv preprint arXiv:2311.09797v2},
  url          = {https://arxiv.org/abs/2311.09797}
}

@inproceedings{lyu2023faithful,
  author    = {Lyu, Qing and Callison-Burch, Chris and others},
  title     = {Faithful Chain-of-Thought Reasoning},
  booktitle = {Proceedings of the 13th International Joint Conference on Natural Language Processing and the 3rd Conference of the Asia-Pacific Chapter of the Association for Computational Linguistics (IJCNLP-AACL 2023)},
  year      = {2023},
  pages     = {305--329},
  publisher = {Association for Computational Linguistics},
  address   = {Nusa Dua, Bali, Indonesia},
  url       = {https://aclanthology.org/2023.ijcnlp-main.20/}
}

@misc{golovneva2023roscoe,
  author       = {Golovneva, Olga and Chen, Moya and Poff, Spencer and Corredor, Martin and Zettlemoyer, Luke and Fazel-Zarandi, Maryam and Celikyilmaz, Asli},
  title        = {{ROSCOE}: A Suite of Metrics for Scoring Step-by-Step Reasoning},
  year         = {2023},
  howpublished = {arXiv preprint arXiv:2212.07919v2},
  url          = {https://doi.org/10.48550/arXiv.2212.07919}
}

@inproceedings{lin2004rouge,
  author    = {Lin, Chin-Yew},
  title     = {{ROUGE}: A Package for Automatic Evaluation of Summaries},
  booktitle = {Text Summarization Branches Out: Proceedings of the ACL-04 Workshop},
  year      = {2004},
  pages     = {74--81},
  publisher = {Association for Computational Linguistics},
  address   = {Barcelona, Spain},
  url       = {https://aclanthology.org/W04-1013/}
}

@inproceedings{zhang2020bertscore,
  author    = {Zhang, Tianyi and Kishore, Varsha and Wu, Felix and Weinberger, Kilian Q. and Artzi, Yoav},
  title     = {BERTScore: Evaluating Text Generation with BERT},
  booktitle = {Proceedings of the International Conference on Learning Representations (ICLR 2020)},
  year      = {2020},
  address   = {Addis Ababa, Ethiopia},
  eprint    = {1904.09675},
  archivePrefix = {arXiv},
  primaryClass  = {cs.CL},
  doi       = {10.48550/arXiv.1904.09675}
}

@inproceedings{reimers2019sentence,
  author    = {Reimers, Nils and Gurevych, Iryna},
  title     = {Sentence-BERT: Sentence Embeddings using Siamese BERT-Networks},
  booktitle = {Proceedings of the 2019 Conference on Empirical Methods in Natural Language Processing and the 9th International Joint Conference on Natural Language Processing (EMNLP-IJCNLP)},
  year      = {2019},
  pages     = {3982--3992},
  publisher = {Association for Computational Linguistics},
  address   = {Hong Kong, China},
  doi       = {10.18653/v1/D19-1410}
}

@misc{grattafiori2024llama,
  author       = {Grattafiori, Aaron and Dubey, Abhimanyu and Jauhri, Abhinav and Pandey, Abhinav and Kadian, Abhishek and Al-Dahle, Ahmad and Letman, Aiesha and Mathur, Akhil and Schelten, Alan and Vaughan, Amy and others},
  title        = {The {Llama 3} Herd of Models},
  year         = {2024},
  howpublished = {arXiv preprint arXiv:2407.21783},
  eprint       = {2407.21783},
  archivePrefix = {arXiv},
  primaryClass  = {cs.AI},
  doi          = {10.48550/arXiv.2407.21783}
}

@article{guo2025deepseek,
  author    = {Guo, Daya and Yang, Dejian and Zhang, Haowei and Song, Junxiao and Zhang, Ruoyu and Xu, Runxin and Zhu, Qihao and Ma, Shirong and Wang, Peiyi and Bi, Xiao and others},
  title     = {{DeepSeek-R1}: Incentivizing Reasoning Capability in LLMs via Reinforcement Learning},
  journal   = {Nature},
  year      = {2025},
  volume    = {645},
  pages     = {633--638},
  eprint    = {2501.12948},
  archivePrefix = {arXiv},
  primaryClass  = {cs.CL},
  doi       = {10.48550/arXiv.2501.12948}
}

@misc{qwen2024,
  author       = {{Qwen Team}},
  title        = {Qwen2.5 Technical Report},
  year         = {2024},
  howpublished = {arXiv preprint arXiv:2412.15115},
  eprint       = {2412.15115},
  archivePrefix = {arXiv},
  primaryClass  = {cs.CL},
  doi          = {10.48550/arXiv.2412.15115}
}

@inproceedings{xie2026clef,
  author    = {Xie, Zhuohan and Elbadry, Rania and Zhang, Fan and Georgiev, Georgi and Peng, Xueqing and Qian, Lingfei and Huang, Jimin and Dimitrov, Dimitar and Jani, Vanshikaa and Dai, Yuyang and others},
  title     = {The {CLEF-2026} {FinMMEval} Lab: Multilingual and Multimodal Evaluation of Financial {AI} Systems},
  booktitle = {Advances in Information Retrieval. ECIR 2026},
  year      = {2026},
  series    = {Lecture Notes in Computer Science},
  volume    = {16486},
  publisher = {Springer, Cham},
  doi       = {10.1007/978-3-032-21321-1_37},
  eprint    = {2602.10886},
  archivePrefix = {arXiv},
  primaryClass  = {cs.CL},
  note      = {Also available as arXiv:2602.10886}
}

@inproceedings{zheng2023judging,
  author    = {Zheng, Lianmin and Chiang, Wei-Lin and Sheng, Ying and Zhuang, Siyuan and Wu, Zhanghao and Zhuang, Yonghao and Lin, Zi and Li, Zhuohan and Li, Dacheng and Xing, Eric P. and others},
  title     = {Judging LLM-as-a-Judge with MT-Bench and Chatbot Arena},
  booktitle = {NeurIPS 2023 Datasets and Benchmarks Track},
  year      = {2023},
  address   = {New Orleans, LA, USA},
  eprint    = {2306.05685},
  archivePrefix = {arXiv},
  primaryClass  = {cs.CL},
  doi       = {10.48550/arXiv.2306.05685}
}

@manual{ollama2025,
  title        = {Ollama: Get up and running with large language models locally},
  author       = {{Ollama Team}},
  year         = {2025},
  note         = {Version 0.5.0, available at \url{https://ollama.com/}},
  url          = {https://ollama.com/}
}

@book{arabov2019excel,
  author    = {Arabov, M. K. and Mamatkulov, A. A. and Solieva, L. F.},
  title     = {Reshenie finansovo-ekonomicheskikh zadach posredstvom {Excel}: Uchebno-metodicheskoe posobie},
  year      = {2019},
  publisher = {RTSU},
  address   = {Dushanbe},
  pages     = {200}
}

@misc{team2025gemma3,
  author       = {{Gemma Team, Google DeepMind}},
  title        = {{Gemma 3}: Improving Open Language Models at a Practical Size},
  year         = {2025},
  howpublished = {arXiv preprint arXiv:2503.19786},
  url          = {https://arxiv.org/abs/2503.19786}
}

@misc{abdin2025phi4mini,
  author       = {Abouelenin, Abdelrahman and Ashfaq, Atabak and Atkinson, Adam and Awadalla, Hany and Bach, Nguyen and Bao, Jianmin and Benhaim, Alon and Cai, Martin and Chaudhary, Vishrav and Chen, Congcong and others},
  title        = {{Phi-4-Mini} Technical Report: Compact yet Powerful Multimodal Language Models via Mixture-of-LoRAs},
  year         = {2025},
  howpublished = {arXiv preprint arXiv:2503.01743},
  eprint       = {2503.01743},
  archivePrefix = {arXiv},
  primaryClass  = {cs.CL},
  url          = {https://arxiv.org/abs/2503.01743}
}

@misc{mistral2024,
  author       = {Jiang, Albert Q. and Sablayrolles, Alexandre and Mensch, Arthur and Bamford, Chris and Chaplot, Devendra Singh and de las Casas, Diego and Bressand, Florian and Lengyel, Gianna and Lample, Guillaume and Saulnier, Lucile and others},
  title        = {Mistral 7B},
  year         = {2023},
  howpublished = {arXiv preprint arXiv:2310.06825},
  eprint       = {2310.06825},
  archivePrefix = {arXiv},
  primaryClass  = {cs.CL},
  url          = {https://arxiv.org/abs/2310.06825}
}

@misc{arya2025ayaexpanse,
  author       = {Dang, John and Singh, Shivalika and D'souza, Daniel and Ahmadian, Arash and Salamanca, Alejandro and Smith, Madeline and Peppin, Aidan and Hong, Sungjin and Govindassamy, Manoj and Zhao, Terrence and others},
  title        = {Aya Expanse: Combining Research Breakthroughs for a New Multilingual Frontier},
  year         = {2024},
  howpublished = {arXiv preprint arXiv:2412.04261},
  eprint       = {2412.04261},
  archivePrefix = {arXiv},
  primaryClass  = {cs.CL},
  url          = {https://arxiv.org/abs/2412.04261}
}

@misc{rusfinchain_hf,
  author       = {Arabov, M. K.},
  title        = {{RusFinChain}: A Symbolic Financial Reasoning Benchmark for Russian LLMs},
  year         = {2026},
  howpublished = {Hugging Face Datasets},
  url          = {https://huggingface.co/datasets/RusNLPWorld/RusFinChain},
  note         = {5,280 tasks, 17 domains, 172 topics}
}

@misc{rusfinchaineval2026,
  author       = {Arabov, M. K.},
  title        = {{RusFinChain-Eval}: Evaluation Results for 8 LLMs on the RusFinChain Benchmark},
  year         = {2026},
  howpublished = {Hugging Face Datasets},
  url          = {https://huggingface.co/datasets/RusNLPWorld/RusFinChain-Eval},
  note         = {8,100 evaluated records, 8 models}
}

\end{document}